\documentclass[sigconf]{acmart}
\acmConference[Accepted at ICSE 2022]{The 44th International Conference on Software Engineering}{May 21–29, 2022}{Pittsburgh, PA, USA}

\usepackage{booktabs}
\usepackage{enumitem}
\usepackage{listings}
\usepackage{xspace}
\usepackage{amsmath}
\usepackage{amsfonts}
\usepackage{graphicx}
\usepackage{textcomp}
\usepackage{xcolor}
\usepackage{algorithm}
\usepackage{algorithmicx}
\usepackage{algpseudocode}

\usepackage{appendix}

\usepackage{cleveref}
\usepackage{acronym}

\usepackage{proof}
\usepackage{subcaption}
\usepackage{tikz}
\usetikzlibrary{positioning,fit,calc}
\usetikzlibrary{tikzmark}

\usepackage[nounderscore]{syntax}
\usepackage{mdframed}
\usepackage{xcolor}
\usepackage{framed}

\setcopyright{none}
\renewcommand\footnotetextcopyrightpermission[1]{} 

\newcommand{\ie}{i.e., }
\newcommand{\eg}{e.g., }

\newcommand{\code}[1]{{\small\texttt{#1}}}

\definecolor{mpcolor}{rgb}{0.8, 0.33, 0.0}

\definecolor{hycolor}{rgb}{0.1,0.6,0.1}

\definecolor{rscolor}{rgb}{0.16,0.32,0.75}

\definecolor{aucolor}{rgb}{0.7,0.11,0.11}

\definecolor{smcolor}{rgb}{0.58,0.0,0.83}

\definecolor{todocolor}{rgb}{1,0.3,0.3}

\definecolor{mauve}{HTML}{E0D7FF}
\definecolor{orange}{HTML}{FFE5B4}
\definecolor{blue}{HTML}{BFEFFF}
\definecolor{light-gray}{gray}{0.80}
\definecolor{lightgreen}{HTML}{CCFFCC}

\acrodef{dag}[DAG]{Directed Acyclic Graph}
\acrodef{fpg}[FPG]{Feature Provenance Graph}

\newcommand{\tool}{\textsc{SapientML}\xspace}
\newcommand{\autosk}{{auto-sklearn}\xspace}
\newcommand{\openml}{{OpenML}\xspace}
\newcommand{\gh}{{GitHub}\xspace}
\newcommand{\mulan}{{Mulan}\xspace}

\newcommand{\baseadapt}{{Baseline1}\xspace}
\newcommand{\baseseeding}{{Baseline2}\xspace}
\newcommand{\basicml}{{Basic-ML}\xspace}
\newcommand{\default}{{Default}\xspace}

\newcommand{\numBM}{41}
\newcommand{\numNewBM}{10}
\newcommand{\numLargeBM}{10}
\newcommand{\numCorpDatasets}{170}
\newcommand{\numCorpNtbks}{1,094}
\newcommand{\numAutoMLComp}{3}

\newcommand{\numBaselines}{4}
\newcommand{\numWinOrComparable}{27}
\newcommand{\numLargeSMLWins}{9}
\newcommand{\numAlFails}{9}
\newcommand{\numLargeAlFails}{4}

\newcommand{\targetColumns}{relevant columns }

\title{\tool: Synthesizing Machine Learning Pipelines by Learning from Human-Written Solutions}

\author{Ripon K. Saha}
\affiliation{
\institution{Fujitsu Research of America, Inc.}
\country{}
}
\email{rsaha@fujitsu.com}

\author{Akira Ura}
\affiliation{
\institution{Fujitsu Ltd.}
\country{}
}
\email{ura.akira@fujitsu.com}

\author{Sonal Mahajan}
\affiliation{
\institution{Fujitsu Research of America, Inc.}
\country{}
}
\email{smahajan@fujitsu.com}

\author{Chenguang Zhu$^\dagger$}\thanks{$^\dagger$ These authors contributed to this work as interns at Fujitsu Research of America.}
\affiliation{
\institution{The University of Texas at Austin}
\country{}
}
\email{cgzhu@utexas.edu}

\author{Linyi Li$^\dagger$}
\affiliation{
\institution{University of Illinois at Urbana-Champaign}
\country{}
}
\email{linyi2@illinois.edu}

\author{Yang Hu$^\dagger$}
\affiliation{
\institution{The University of Texas at Austin}
\country{}
}
\email{huyang@utexas.edu}

\author{Hiroaki Yoshida}
\affiliation{
\institution{Fujitsu Research of America, Inc.}
\country{}
}
\email{hyoshida@fujitsu.com}

\author{Sarfraz Khurshid}
\affiliation{
\institution{The University of Texas at Austin}
\country{}
}
\email{khurshid@utexas.edu}

\author{Mukul R. Prasad}
\affiliation{
\institution{Fujitsu Research of America, Inc.}
\country{}
}
\email{mukul@fujitsu.com}

\begin{document}

\begin{abstract}
Automatic machine learning, or AutoML, holds the promise of truly democratizing the use of machine learning (ML), by substantially automating the work of data scientists. However, the huge combinatorial search space of candidate pipelines means that current AutoML techniques, generate sub-optimal pipelines, or none at all, especially on large, complex datasets. In this work we propose an AutoML technique \tool, that can learn from a corpus of existing datasets and their human-written pipelines, and efficiently generate a high-quality pipeline for a predictive task on a new dataset. To combat the search space explosion of AutoML, \tool employs a novel divide-and-conquer strategy realized as a three-stage program synthesis approach, that reasons on successively smaller search spaces. The first stage uses meta-learning to predict a set of plausible ML components to constitute a pipeline. In the second stage, this is then refined into a small pool of viable concrete pipelines using a pipeline dataflow model derived from the corpus. Dynamically evaluating these few pipelines, in the third stage, provides the best solution. We instantiate \tool\ as part of a fully automated tool-chain that creates a cleaned, labeled learning corpus by mining Kaggle, learns from it, and uses the learned models to then synthesize pipelines for new predictive tasks. We have created a training corpus of \numCorpNtbks\ pipelines spanning \numCorpDatasets\ datasets, and evaluated \tool\ on a set of \numBM\ benchmark datasets, including \numNewBM\ new, large, real-world datasets from Kaggle, and against \numAutoMLComp\ state-of-the-art AutoML tools and \numBaselines\ baselines. Our evaluation shows that \tool{} produces the best or comparable accuracy on \numWinOrComparable\ of the benchmarks while the second best tool fails to even produce a pipeline on \numAlFails\ of the instances. This difference is amplified on the \numLargeBM\ most challenging  benchmarks, where \tool{} wins on \numLargeSMLWins\ instances with the other tools failing to produce pipelines on \numLargeAlFails\ or more benchmarks.
\end{abstract}



\maketitle

\section{Introduction}
\label{sec:introduction}

The explosive growth in machine learning (ML) applications, over the past decade, has created a huge demand for data scientists (DS) and ML practitioners to develop real-world ML solutions. The 2018 LinkedIn Workforce Report showed a shortage of 151,717 DS, nationwide~\cite{linkedIn-report2018}, that had grown to 250,000 by 2020~\cite{QuantHub-2020}.  Automatic machine learning, or AutoML, holds the promise of addressing this shortfall~\cite{automl_book, automl-survey2018, automl-survey2019}. AutoML can improve productivity of data science teams and cover gaps in expertise.

Given a dataset and a predictive task (\eg classification or regression) AutoML aims to create an ML pipeline that trains an optimized ML model for the given task. Simply put, the pipeline is a sequence of ML operators that processes data to make it suitable for learning (feature engineering (FE)), fits a suitable ML model on it (model selection), and calculates the predictive performance of the model. One of the prominent instances of AutoML, the subject of much research recently, is creating supervised ML pipelines for tabular data~\cite{TPOT:GECCO2016, Auto-sklearn-NIPS2015, AL:OOPSLA2019,  xudong2019reinbo, AMS:FSE2020, OBOE-KDD2019, TensorOBOE-KDD2020}. This paper also focuses on this formulation of AutoML.

AutoML has been traditionally solved as a search and optimization problem -- \textit{selecting} the best pipeline from a space of candidates~\cite{TPOT:GECCO2016, Auto-WEKA-KDD2013, Auto-sklearn-NIPS2015, OBOE-KDD2019, TensorOBOE-KDD2020}. However, ML pipelines are also programs, in fact relatively small, highly structured domain-specific programs, that could be amenable to program synthesis. Further, public repositories like Kaggle~\cite{kaggle} and \gh contain hundreds of thousands of human-written ML pipelines that could serve as starting points for synthesizing new pipelines. Indeed, program synthesis by mining or learning from existing program corpora has been successfully deployed for other endpoints of synthesis~\cite{Raychev2014, Lee2018, murali2018neural, CoCoNuTISSTA2020, GetAFix:OOPSLA:2019}. Emerging research~\cite{AL:OOPSLA2019, AMS:FSE2020} demonstrates the promise of this perspective for ML pipeline synthesis. Our work also follows this philosophy but offers a novel take on the core challenge of AutoML. 

The central challenge of AutoML is the massive \textit{combinatorial} search space of candidate ML pipelines it exposes -- compositions of different potential FE operators, each applied on different columns of the subject dataset, composed with one of several potential models or their ensembles. Further, each pipeline component may have its own space of hyper-parameters. Previous AutoML techniques adopt several approaches to combat this combinatorial explosion.
Some try to search a restricted search space by excluding FE from consideration~\cite{Auto-WEKA-KDD2013, Auto-sklearn-NIPS2015, OBOE-KDD2019}, searching specific pipeline topologies~\cite{TPOT:GECCO2016}, or a pre-compiled explicit corpus of synthetic pipelines~\cite{Auto-sklearn-NIPS2015, TensorOBOE-KDD2020, OBOE-KDD2019}. Others try to prune the search space by using learned language models coupled with aggressive dynamic evaluation of \textit{partial pipelines}~\cite{AL:OOPSLA2019} or by warm-starting search using constraints mined from human-written pipelines~\cite{AMS:FSE2020}. 
However, navigating the huge combinatorial search space of AutoML remains an open problem. In fact, \textit{none} of the above techniques  apply FE transforms to specific dataset columns, as a human DS would. Instead, they are blindly applied to the complete dataset, ostensibly to avoid injecting another set of hyper-parameters into the search space.

\textbf{Insight.} Our key insight is that the root cause of the AutoML search-space explosion is because previous AutoML techniques reason on complete ML pipelines (\textit{combinations} of various ML components) as a single entity, ostensibly to capture dependencies between ML components. However, we observe that in many practical instances, the decision on whether to include a particular component (say an \textit{imputer}) in a pipeline can be made based primarily on properties of the dataset (whether or not it has missing values), \textit{independent} of other components. Indeed, human DS often employ such \textit{best practices} when manually constructing ML pipelines. Once the set of \textit{plausible components} to use for a given dataset are known, they can be used to assemble a target pipeline or a small population of plausible target pipelines to choose from. This would substantially mitigate the combinatorial explosion coming from exploring arbitrary combinations of components. Further, we hypothesize that these DS best practices are represented in publicly available human-written ML pipelines (say on Kaggle). Thus, these pipelines can potentially be mined to learn and then replicate human DS decision making to create viable pipelines for new datasets.

\textbf{Proposed approach.} Pursuant to the above insight, we propose an AutoML technique \tool~\footnote{an AutoML approach harnessing the wisdom (\textit{sapere}) of human (\textit{sapien}) data scientists.}, that can learn from a corpus of existing datasets and their pipelines, and generate a high-quality pipeline for a predictive task on a new dataset.
To combat the search space explosion of AutoML, \tool employs a novel divide-and-conquer strategy, realized as a three-stage program synthesis approach that reasons on successively smaller search spaces. The first stage uses meta-learning to train a meta-model (offline phase) which is then used to \emph{independently} predict the \textit{suitability} of each ML component with respect to the given dataset (in the online phase). Specifically, this meta-model captures the relationship between features of the dataset (\eg the presence of missing data values) and desired components in the pipeline (\eg a data interpolation component). This prediction yields a ranked list of \textit{pipeline skeletons}. Each pipeline skeleton is an (unordered) set of plausible ML components, to constitute a pipeline, each component mapped to specific (or all) dataset columns on which it should be applied. In the second stage, the skeletons are then \textit{instantiated} into a small pool of viable concrete pipelines using a pipeline \emph{dataflow model} mined from the corpus, and a small library of standard implementations for each ML component. For each candidate skeleton the pipeline components are correctly ordered and incompatible components discarded using the dataflow model, and each component instantiated using code templates from the library. Dynamically evaluating these few pipelines (the most expensive operation), in the third stage, yields the best pipeline. The concept of a multi-stage approach employing more expensive analyses on successively smaller spaces has been successfully used in other domains, including automatic program repair~\cite{CoCoNuTISSTA2020} and code search~\cite{AROMA:OOPSLA2019}, among others. Our specific design is customized for ML pipeline synthesis.

We instantiate \tool as part of a fully automatic end-to-end tool-chain that mines datasets and corresponding pipelines from Kaggle, automatically cleans and labels each pipeline, learns from this corpus and then synthesizes ML pipelines for predictive tasks on new datasets. We evaluate \tool on a set of \numBM\ benchmark datasets, including \numNewBM\ new, large, real-world datasets from Kaggle and against \numAutoMLComp\ state of the art AutoML tools (AL~\cite{AL:OOPSLA2019}, \autosk~\cite{Auto-sklearn-NIPS2015}, TPOT~\cite{TPOT:GECCO2016}) and \numBaselines\ baselines. Our evaluation shows that \tool{} produces the best or comparable accuracy on \numWinOrComparable\ of the benchmarks while the second best tool (AL), fails to even produce a pipeline on \numAlFails\ of the instances. Further, on the most challenging \numLargeBM\ benchmarks \tool{} wins on \numLargeSMLWins\ instances with the other AutoML tools failing on \numLargeAlFails\ or more benchmarks. 

This paper makes the following main contributions: 
\begin{itemize}[leftmargin=3mm,topsep=0.5mm,parsep=0mm]
    \item \textbf{Technique:} A learning-based AutoML technique \tool, that can efficiently synthesize high-quality supervised ML pipelines, using a novel divide-and-conquer approach to circumvent the combinatorial state-space explosion of AutoML.
    \item \textbf{Tool:} An implementation of \tool\ as part of an automated tool-chain that creates a cleaned, labeled learning corpus by mining Kaggle, learns from it, and uses the learned models to then synthesize pipelines for predictive tasks on new datasets. 
    \item \textbf{Evaluation:} A substantial evaluation of \tool\ on a benchmark of \numBM\ datasets, including \numNewBM\ new, large, real-world datasets from Kaggle, comparing it to \numAutoMLComp\ state of the art AutoML tools and \numBaselines\ baseline techniques for creating ML pipelines.
\end{itemize}

\section{Problem Definition}

A \textit{tabular dataset}, $D=(X\times Y) \in \mathcal{D}$ is sampled from a distribution over a domain $\mathcal{X}\times\mathcal{Y}$ where $\mathcal{X}$ and $\mathcal{Y}$ denote an input domain and an output domain respectively. $X$ is comprised of $n$ rows and $d$ columns, called \textit{features}, where each row represents an observation consisting of $d$ values from $\mathcal{X}$. Similarly $Y$ is comprised of $n$ rows and $t$ columns where each row is a $t$-tuple of values or labels from $\mathcal{Y}$. A \textit{supervised predictive task} on $D$ is to learn a prediction function $h : X \xrightarrow{} Y$ such that $y \approx h(x)$. A predictive task is called a {\it classification} task when the $y$ is discrete and called a {\it regression} task when $y$ is continuous. For multi-label classification and multivariate regression, $|t|>1$. Applying supervised machine learning (ML) to a predictive task requires a training version of the dataset $D_{train}$ to train an ML model, and a held out test dataset, $D_{test}$ to evaluate its performance. A single dataset $D$ can also be split into $D_{train}$ and $D_{test}$. 

Given a dataset $D$, an \textit{ML pipeline} ($P \in \mathcal{P}$) is a sequence of FE components followed by a model component that realizes a given predictive task. Hence, $P =  [ c^1_f, c^2_f,..,c^k_f, c_m ]$ represents a pipeline with $k$ FE components and one model. 
A pipeline \textit{component} $c \in \mathcal{C}$ is comprised of one or more API calls, and associated glue code, that together performs an atomic data-specific pipeline task, \eg filling missing values or transforming a categorical column to a set of numeric columns. There are two kinds of components: i) the FE components ($c_f$) that transforms a feature ($x$) or a set of features ($X' \subset X$) including data wrangling tasks, and ii) the model components ($c_m$) that performs the actual learning and prediction. 

Given dataset $D = D_{train} \cup D_{test}$, a predictive task on $D$, and an accuracy metric $\sigma$ (\eg F1 score for classification and $R^2$ for regression problems respectively), our aim is to create an executable machine learning pipeline $P$ for this dataset and task that maximizes (without loss of generality) $\sigma$ on $D_{test}$. We pose this \textit{pipeline synthesis} problem as a program synthesis problem with quantitative objectives, akin to \cite{QSyGuS-CAV2018}.

\section{Motivating Example}
\label{sec:motivating-example}

\begin{figure}[t]
\captionsetup{justification=centering}
\centering
    \begin{subfigure}[b]{0.45\textwidth}
    \centering
        \begin{tabular}{|l @{\hspace{2em}} r|}
            \hline
            $\langle$\code{FE:OrdinalEncoder}(\code{card4, \ldots}),  0.73$\rangle$ & {\large \textbf{$C'_{f}$}}\\
            $\langle$\code{FE:OneHotEncoder}(\code{card4, \ldots}), 0.70 $\rangle$ &\\
            $\langle$\code{FE:Imputer}(\code{card2, card3,\ldots}), 0.81$\rangle$ &\\
            $\langle$\code{FE:LinearScaler}($X$)$, 0.69\rangle$ &\\
            $\langle$\code{FE:DataBalancer}($X$), 0.58 $\rangle$ &\\
            \hline\hline
            $\langle$\code{MODEL:CatBoostClassifier}($X$), 1$\rangle$ &\\
            $\langle$\code{MODEL:ExtraTreesClassifier}($X$), 2$\rangle$  & {\large \textbf{$C'_{m}$}}\\
            $\langle$\code{MODEL:XGBClassifier}($X$), 3$\rangle$&\\
            
            \hline
        \end{tabular}
     \caption{\label{fig:predicted_components}Predicted FE and Model components by the skeleton predictor with their probability scores and rank respectively}

     \centering
        \begin{tabular}{|l @{\hspace{2em}}|}
            \hline
            $C'_f ~\cup $~$\langle$\code{MODEL:CatBoostClassifier}($X$)$\rangle$\\\hline\hline
            $C'_f ~\cup $~$\langle$\code{MODEL:ExtraTreesClassifier}($X$)$\rangle$\\\hline\hline
            $C'_f ~\cup $~$\langle$\code{MODEL:XGBClassifier}($X$)$\rangle$\\
            \hline
        \end{tabular}
     \caption{\label{fig:skeleton}Three skeletons generated by the pipeline seeding phase} 
     
     \begin{tabular}{|l @{\hspace{2em}}|}
            \hline
            $\langle$\code{FE:Imputer}(\code{card2, card3,\ldots})$\rangle$\\
            $\langle$\code{FE:OrdinalEncoder}(\code{card4, \ldots})$\rangle$\\
            $\langle$\code{FE:LinearScaler}($X$)$\rangle$\\
            $\langle$\code{FE:DataBalancer}($X$) $\rangle$\\
            $\langle$\code{MODEL:CatBoostClassifier}($X$)$\rangle$ \\
            \hline
        \end{tabular}
     \caption{\label{fig:ordered_skeleton}First skeleton after ordering and redundancy removal} 
    \end{subfigure}
    
    \caption{\label{fig:example}Artifacts of Pipeline Seeding and Pipeline Instantiation Phases for IEEE-CIS-Fraud-Detection Example}
\end{figure}

\begin{figure*}[t]
\begin{mdframed}
\setlength\fboxsep{0mm}
\begin{lstlisting}[language=Java, mathescape, framexleftmargin=0em, basicstyle=\fontsize{7pt}{7pt}\selectfont\ttfamily, numberstyle=\scriptsize, numbers=none]
import pandas as pd @\hspace{1em}\colorbox{light-gray}{\strut \# LOAD DATA}@
__train_dataset=pd.read_csv("training.csv", delimiter=",") 
__test_dataset=pd.read_csv("test.csv", delimiter=",")

from sklearn.impute import SimpleImputer @\hspace{1em}\colorbox{orange}{\strut \# FE TRANSFORM 1}@
import numpy as np
_NUMERIC_COLS_WITH_MISSING_VALUES = ['card2', 'card3', .. 'V339']
for _col in _NUMERIC_COLS_WITH_MISSING_VALUES:
    __imputer = SimpleImputer(missing_values=np.nan, strategy='mean')
    __train_dataset[_col] = __imputer.fit_transform(__train_dataset[_col].values.reshape(-1,1))[:,0]
    __test_dataset[_col] = __imputer.transform(__test_dataset[_col].astype(
                                                    __train_dataset[_col].dtypes).values.reshape(-1,1))[:,0]
                                                    
_STRING_COLS_WITH_MISSING_VALUES = ['card4', 'card6',..., 'M9'] @\hspace{1em}\colorbox{orange}{\strut \# FE TRANSFORM 2}@
@{\Large $\dots$}@  ## Apply SimpleImputer for string columns similar to the FE transform 1 shown above 

from sklearn.preprocessing import OrdinalEncoder @\hspace{1em}\colorbox{orange}{\strut \# FE TRANSFORM 3}@
@{\Large $\dots$}@  ## Apply OrdinalEncoder to categorical columns ['ProductCD', 'card4', .., 'M9']

__feature_train = __train_dataset.drop(['isFraud'], axis=1) @\hspace{1em}\colorbox{light-gray}{\strut \# DETACH TARGET}@
__target_train =__train_dataset['isFraud']
__feature_test,  __target_test = __test_dataset.drop(['isFraud'], axis=1), __test_dataset['isFraud']

from sklearn.preprocessing import StandardScaler @\hspace{1em}\colorbox{orange}{\strut \# FE TRANSFORM 4}@
@{\Large $\dots$}@  ## Apply StandardScaler to __feature_train and __feature_test

from imblearn.over_sampling import SMOTE @\hspace{1em}\colorbox{orange}{\strut \# FE TRANSFORM 5}@
__feature_train, __target_train = SMOTE().fit_resample(__feature_train, __target_train)

from catboost import CatBoostClassifier @\hspace{1em}\colorbox{lightgreen}{\strut \# MODEL}@
__model = CatBoostClassifier()
__model.fit(__feature_train, __target_train)
__y_pred = __model.predict(__feature_test)

from sklearn import metrics @\hspace{1em}\colorbox{light-gray}{\strut \# EVALUATION}@
print(metrics.f1_score(__target_test, __y_pred, average='macro'))
\end{lstlisting}
\end{mdframed}
\caption{Abridged version of pipeline generated by \tool{} for the IEEE-CIS-Fraud-Detection prediction task}
\label{fig:IEE_CIS_pipeline}
\end{figure*}

In this section, we illustrate the use-case and mechanics of our technique using a real-world dataset \emph{IEEE CIS Fraud Detection}~\cite{ieeecis}, provided by the company Vesta and hosted on Kaggle. It contains 591K rows of data, each corresponding to an e-commerce transaction represented by a rich set of 394 features. The features are mainly numeric (\eg transaction amount) and string categorical values (\eg device type). Some features are missing in some transactions. 
The predictive task is to label a transaction as fraudulent or not, based on its features, \ie a binary classification task.  

\textbf{Use Case.} Creating a pipeline for a predictive task may take a long time for a DS. The DS needs to decide on the appropriate set of feature engineering (FE) ($c_f$) and model ($c_m$) components to use, the right dataset columns (features) to apply each of them on, and then instantiate them in the right order so the pipeline executes on the dataset ($D$) without errors. Given the huge space of possibilities for these decisions, data scientists typically rely on their understanding of $D$, past experience, and often brute-force trial and error, to complete this laborious task~\cite{khurana2018feature}. AutoML tools can accelerate this process substantially, especially for a novice DS, by providing her with a good-quality, executable pipeline for potential last-mile optimization.

\textbf{Key Challenge.} Real-world, large, complex datasets like IEEE CIS Fraud Detection present particularly challenging cases for current AutoML tools. In order to navigate the huge combinatorial search space of possible candidate pipelines tools such as TPOT~\cite{TPOT:GECCO2016} and \autosk~\cite{Auto-sklearn-NIPS2015} restrict themselves to numeric data, which ensures smaller, simpler pipelines. Thus, they cannot even run on the given dataset since it has string categorical features. The state-of-the-art tool AL~\cite{AL:OOPSLA2019} uses a combination of learned language model and aggressive dynamic evaluation of partial pipelines to search for a viable solution. However, in this case, it evaluates 1,641 partial and 1,310 complete pipelines in 1.5 hours (on a 8 vCPU and 32GB memory machine) and finally crashes due to an internal timeout without producing any pipeline.

{\bf \tool's} three-stage program synthesis approach proves to be quite effective on this example. 
In the first stage (Section~\ref{sec:pipeline-seeding}), \tool{} uses a machine-learned model, trained on its \textit{meta-learning corpus} of human-written pipelines, to generate a ranked-list of \textit{pipeline skeletons}, to construct viable pipelines. 
For the present example, \tool{} first predicts five potential FE components and the top three most appropriate models in \Cref{fig:predicted_components} to generate three pipeline skeletons in \Cref{fig:skeleton}. The predicted components broadly agree with human intuition. For instance, \code{OrdinalEncoder} and \code{OneHotEncoder} are reasonable transforms to encode the String-based features in the dataset and the use of \code{DataBalancer} comports with the significant imbalance between the number of fraudulent and valid transactions in the dataset. Further, the choice of the \code{CatBoost} model is consistent with the previous research~\cite{tanha2020boosting} showing that \code{CatBoost} performs well for classification on large, imbalanced data.
\tool{} further uses the decision rules learned by the skeleton predictor to infer the relevant features in the dataset where each FE transform will be applied. For example, \Cref{fig:predicted_components} shows that \tool{} targets \code{card2, \ldots} for \code{SimpleImputer}.

\sloppy

In the second stage (Section~\ref{sec:pipeline-instantiation}), \tool concretizes the pipeline skeleton into a set of executable pipelines. To this end, it uses the confidence scores included in the skeleton as well as a \textit{pipeline dataflow meta-model} mined (offline) from the learning corpus to discard redundant FE components, and order the components in a syntactically correct fashion, to produce ordered skeletons, as shown in Figure~\ref{fig:ordered_skeleton}.  For instance, the analysis concludes that both \code{OrdinalEncoder} and \code{OneHotEncoder} are to be applied on the same dataset columns but cannot be simultaneously used. Thus, \code{OneHotEncoder} which has a lower confidence score, is discarded. As another example, \code{Imputer} is ordered before \code{OrdinalEncoder}, following the mined component order relation. 
Next the ordered skeletons are transformed into a set of concrete pipelines (three in this case).

In the final stage, \tool evaluates these candidate pipelines on a held-out validation dataset (derived from only the training dataset, \textit{not} the testing dataset) and returns the highest accuracy pipeline. Figure~\ref{fig:IEE_CIS_pipeline} shows an abridged version of this pipeline.
The pipeline implements a rich set of five FE components each applied to its appropriate columns and paired with a \code{CatBoost} gradient-boosting classification model. \tool takes only 8 mins to produce this pipeline and produces a respectable 0.82 F1 score. 

\section{Approach}
\label{sec:approach}

\subsection{Overview}

\begin{figure*}[t]
\begin{center}
\includegraphics[width=.95\linewidth]{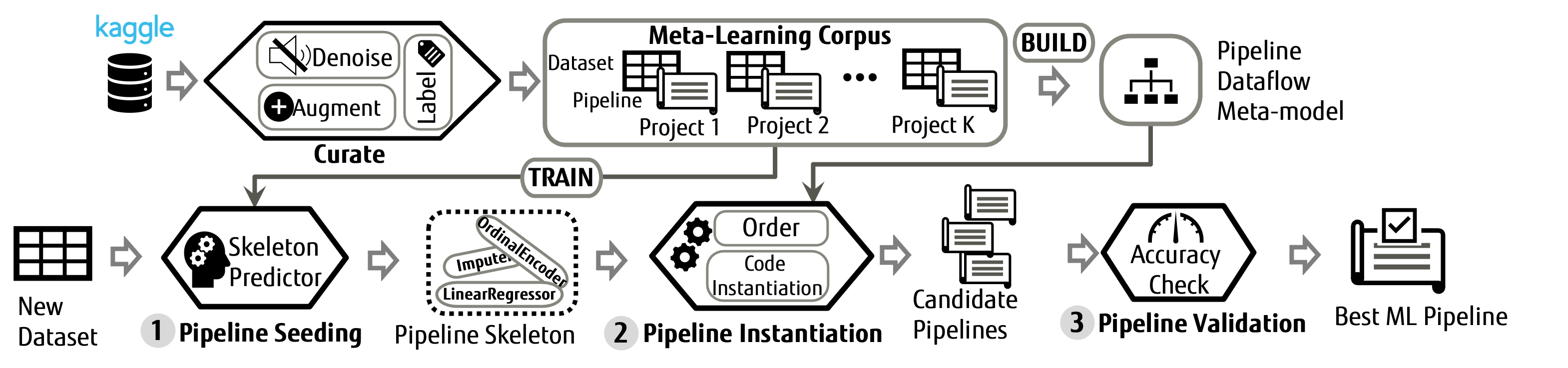}
\end{center}
\caption{Overview of \tool{} system.}
\label{fig:overview}
\end{figure*}

Figure~\ref{fig:overview} presents a high-level overview of the \tool{} system. It has an offline and an online phase. In the offline phase, \tool{} creates a corpus of human-written pipelines and their datasets, called the \textit{meta-learning corpus}, by mining data-science repositories (Kaggle in our case) and automatically curating the data for learning, through denoising, augmentation, and labeling. The meta-learning corpus is then used to build two meta-models, namely the \textit{skeleton predictor} meta-model and the \textit{pipeline dataflow meta-model}. In the online phase, given a new dataset and a predictive task (classification or regression) defined on it, \tool{} uses two meta-models to synthesize a supervised ML pipeline for the given dataset and task, which maximizes some performance metric (\eg F1 or R2).

\tool navigates the huge combinatorial search space of AutoML through a novel three-stage program synthesis approach that reasons on successively smaller search spaces. The first stage, called \textit{pipeline seeding}, uses the skeleton predictor derived through meta-learning on the meta-corpus, to independently predict the \textit{suitability} of each ML component to appear in an ML pipeline for the given dataset, based on the meta-features of the dataset. This prediction yields a \textit{pipeline skeleton}, an unordered set of plausible components, to constitute a solution pipeline. In the second stage, called \textit{pipeline instantiation}, this skeleton is concretized into a small pool of viable \textit{candidate pipelines}, using the dataflow meta-model  mined from the corpus, to correctly order, minimize, and instantiate the pipeline components. The final, \textit{pipeline validation} stage selects the highest accuracy ML pipeline among the candidate pipelines by dynamically evaluating them. The following sub-sections describe the meta-corpus creation and the three pipeline synthesis stages. 

\subsection{Creation of the Meta-Learning Corpus}
\label{sec:corpus}

This step automatically mines and curates a high-quality corpus that includes human-written ML pipelines and their datasets, to power the meta-learning of \tool{}. These pipelines naturally capture the expertise and domain knowledge of human DS as opposed to creating a relatively small, homogeneous, synthetic ML pipeline corpora used by some other AutoML techniques~\cite{TensorOBOE-KDD2020, Auto-sklearn-NIPS2015} that incurs significant computational cost. To build the corpus, we first mine the datasets and their pipelines from Kaggle~\cite{metakaggle} -- a popular data-science repository. Specifically,  we collected top 350 datasets based on user votes, and up to 100 top-voted pipelines per dataset, giving us around 2,500 initial pipelines. These raw pipelines and datasets are further denoised, augmented, and labelled to make them suitable for learning by \tool{}. Our final corpus is comprised of \numCorpNtbks\ pipelines across \numCorpDatasets\ datasets.

\subsubsection{Denoising pipelines}
\label{sec:denoise-pipelines}
Human-written notebooks on Kaggle often contain {\it noise} in the form of exploratory data analysis, visualization, and debugging code that while useful for human comprehension, are irrelevant to ML model execution. Further, some pipelines may no longer be executable due to various issues such as deprecated APIs and differences in the run time environment.  To construct a clean meta-learning corpus, we first discard any pipeline that fails to run successfully on our environment. Then to remove the noise in each executable pipeline, $P$, we first heuristically identify a criteria line ($l_{cr}$), which performs the final prediction task. In pipelines using the popular python ML libraries such as Scikit-learn\cite{scikit-learn} and XGBoost~\cite{xgboost}, $l_{cr}$ is typically a call to the \code{predict} API function. Then, we compute a forward slice $P_{forw}$ and a backward slice $P_{back}$ from $l_{cr}$, by applying standard dynamic program slicing~\cite{agrawal1990dynamic} and concatenate $P_{forw}$, $l_{cr}$, and $P_{back}$ to yield the clean pipeline $P_{clean}$. We compare the accuracy scores of $P$ and $P_{clean}$ and discard $P_{clean}$ if it obtains a lower score than $P$.

\subsubsection{Augmenting pipelines}
\label{sec:corpus-augmentation}
This step is motivated by the observation that human-written pipelines may not contain the best representative ML model choice for each dataset. This could happen in pipelines written by novice data scientists or because of the availability of newer, better models after the pipeline was written.
Presence of sub-optimal ML models in our meta-learning corpus in turn degrades the quality of the pipelines synthesized by \tool{}. 
To alleviate this problem we employ a \textit{data augmentation} technique to systematically replace sub-optimal models in meta-corpus pipelines with better, \ie higher-accuracy, models. Data augmentation~\cite{shorten2019survey} is commonly employed in machine learning flows to improve the predictive quality of training data.

{\bf Generation of candidates.} To improve the performance score of a denoised pipeline $P_{clean}$ with model $c_m$, our data augmentation technique systematically replaces the model $c_m$ in $P_{clean}$ by each viable model in the corpus, $\mathcal{C}_m=\{c^1_m,\ldots,c^b_m\}$ {\it one at a time}, to create a set of candidate pipelines $\mathcal{P}_{mutated}=\{P^1_{clean} \ldots P^b_{clean}\}$. To identify the code for $c_m$, we start with $l_{cr}$ (defined in \Cref{sec:corpus-augmentation}) and compute a backward slice up to the model declaration. Next we identify the variable names of the model object, $X_{train},Y_{train}$, $X_{test}$, and $Y_{test}$ through static analysis. Finally we replace the declaration of the old model $c_m$ with a new model $c^i_m$ to generate $P^i_{clean}$.

{\bf Selection.} Each mutated pipeline in $\mathcal{P}_{mutated}$ is run on the corresponding dataset, and the best mutated pipeline, $P^{best}_{clean}$, showing the highest improvement in the performance score replaces the original pipeline in the corpus.

\subsubsection{Creation of abstract pipelines for meta-learning}
\label{sec:auto-labeling}

The objective of creating a meta-learning corpus is to provide \tool{} necessary training data to learn the relationship between various dataset properties and ML components. However, since human-written ML pipelines are diverse in terms of implementation, it is challenging to learn the relationships between various dataset properties and raw code snippets. To keep the meta-learning tractable, we represent a pipeline at an abstract level as a sequence of ML components, $P = [c^1_f, c^2_f,..,c^k_f, c_m]$ where $c^i_f$ and $c_m$ represent the labels of FE and model components respectively. 

To this end, first we automatically annotate each component with a label that has two pieces of information: $\langle$\textit{component\_type}:\textit{API\_Name}$\rangle$.  We primarily distinguish two types of components: feature engineering (\code{FE}) and \code{MODEL}. The automatic labeling process involves two steps: i) extracting the API name, and ii) identifying whether a particular API is an \code{FE} or a \code{MODEL} component. We perform an AST analysis to extract API names from each statement, ignoring any APIs that are part of boilerplate code. For example, almost every FE engineering APIs in Scikit-learn library are accompanied by a template code that contains \code{fit}, \code{transform}, or \code{fit\_transform} APIs. Once we have all the API names, we first annotate the model component (already identified in \Cref{sec:corpus-augmentation} for each pipeline in our corpus). All components appearing before the MODEL component are labeled as FE components.

At this point, the abstract pipeline $P$ is presented at the API level. However, there are many labels in the corpus that are functionally similar across pipelines. For example, a data scientist can use either the \code{fillna} API from \code{pandas} or \code{SimpleImputer} from \code{sklearn} to fill out the missing values in a dataset. To learn meaningful patterns of meta-features with respect to these labels, we have to  group the components that are semantically similar. Applying domain-knowledge in ML is a standard practice and meta-learning is no exception. Therefore, we investigated the labels in our project corpus and group them based on their functionality by studying the API documentation. Then we assigned each group a functional label. For example, we grouped \code{FE:fillna}, \code{FE:interpolate}, \code{FE:SimpleImputer} and \code{FE:KNNImputer} together and mapped to a unified label \code{FE:Imputer} since they all are used to impute missing values in a dataframe. 

\subsection{Stage 1: Pipeline Seeding} 
\label{sec:pipeline-seeding}

Given a dataset ($D$) and predictive task, the objective of the pipeline seeding stage is to produce a ranked-list of {\it pipeline skeletons}, $\mathcal{S}=[S_1,\ldots S_k]$. This is used to constitute concrete candidate pipelines in the subsequent pipeline instantiation stage (\Cref{sec:pipeline-instantiation}).

A pipeline skeleton ($S$) is a (unordered) set  of plausible ML components that includes zero or more FE components and one model component (\Cref{def:skeleton}). 
To predict the ML components in $S$, \tool{} uses a meta-learning model, called the {\it skeleton predictor}, trained during the offline meta-training phase. The skeleton predictor is architected as a set of sub-models, each of which learns a mapping between properties ({\it meta-features}) of a dataset $D$ and the likelihood of a specific ML component ({\it meta-target}) appearing in a pipeline for $D$.

\begin{definition}[Meta-features]
\label{def:meta-features}
A set of meta-features, $\Phi=\{ \alpha_1, \alpha_2, \ldots, \alpha_l \}$, quantify the characteristics of a dataset where each meta-feature is computed by a function $\alpha_i : D \xrightarrow{} \mathbb{R}$ that takes the dataset as input and outputs a real number.
\end{definition}

\subsubsection{Design of Meta-Features} 
A good set of meta-features should have three properties: i) they are expressive enough to characterize the dataset, ii) they are succinct enough so that there exist some meaningful patterns that skeleton predictor can learn with respect to the ML components, and iii) they are efficient to compute. For example, the choice of \code{FE} and \code{MODEL} component often depends on the meta-features such as number of records and features in $D$ and feature types. Based on the existing literature~\cite{Auto-sklearn-NIPS2015, AL:OOPSLA2019} and our experience, we compute 38 meta-features to characterize the datasets in our meta-training corpus.  \Cref{tbl:meta-features} presents the list of meta-features used in \tool{}.

\begin{table}[t]
 \centering
 \footnotesize
 \caption{Summary of Meta-Features}
 \label{tbl:meta-features}
 \begin{tabular}{p{2.8cm}p{5cm}}
 \hline
 High-Level Property & Meta-features \\
 \hline \hline
Shape of dataset (3) & Number of rows, features, and targets\\
Missing entries (1) & Presence of missing values\\
Feature types (10)  & Presence and \#features, whose data type is numeric, number category, string category, text, and date.\\
Measure of symmetry (4) & Skewness and Kurtosis (normal, uniform, and tailed)\\
Measure of Distribution (6) & Normal, Uniform, and Poisson distribution for features and target\\
Tendency and Dispersion (3) & Normalized mean, standard deviation, variation across columns\\
Correlated features (3) & Pearson correlation (min, max, number of correlated features)\\
Outliers (2) & \#features that contains few or many outliers.\\
Value frequency (3) & Number of features whose values are sparse, imbalanced, dominant\\
Target property (3) & Imbalanced, continuous or categorical.\\
\hline
 \end{tabular}
\end{table}

\begin{definition}[Meta-Targets]
\label{def:meta-target}
A set of ML components $\mathcal{C} = \mathcal{C}_f \cup \mathcal{C}_m$ that define the prediction space of skeleton predictor where $\mathcal{C}_f$ and $\mathcal{C}_m$ represent FE and model components respectively. 
\end{definition}

\subsubsection{Meta-targets}
Each ML component in the abstract pipelines created in \Cref{sec:auto-labeling} is a valid meta-target for \tool{}. However, to learn any meaningful pattern between meta-features ($\Phi)$ and a particular ML component $c$, we need sufficient occurrences of $c$ in the meta-corpus. Therefore, we excluded any ML components that appeared less than five times in our corpus. This filtering criteria provided us 9 FE components and 29 model components (15 classification models and 14 regression models) as meta-targets.  Table~\ref{tbl:components} summarizes the ML components that \tool{}'s meta-model predicts.

\begin{table}[t]
 \centering
 \footnotesize
 \caption{Meta-Targets(C: Classification, R: Regression)}
 \label{tbl:components}
 \begin{tabular}{l|lcc|lcc}
 \hline
 Feature Engg. & Model & C & R &  Model & C & R \\
 \hline \hline
Imputer & RandomForest & x & x & SVM & x & x\\
OrdinalEncoder & ExtraTrees & x& x & LinearSVM & x & x\\
OneHotEncoder& LightGBM & x & x & LogisticRegression & x & x\\
TextVectorizer & XGBoost & x & x & Lasso & - & x\\
TextPreprocessor & CatBoost & x & x & SGD & x & x\\
DateFeaturization & GradientBoosting & x & x & MLP & x & x\\
LinearScaler & AdaBoost & x & x & MultinomialNB & x & -\\
LogScaler & DecisionTree  & x & x & GaussianNB & x & -\\
DataBalancer & - & - & - & - & - & -\\
\hline
 \end{tabular}
\end{table}

\subsubsection{Design of Skeleton Predictor (Meta-Models)}

Given a set of meta-features, $\Phi$ computed from $D$, the objective of skeleton predictor is to predict a set of plausible FE components and model components to generate pipeline skeletons defined in \Cref{def:skeleton}. 

\begin{definition}[Skeleton]
\label{def:skeleton}
A skeleton, $S=\{\langle c_f^1(X_1),\rho_1\rangle, \ldots, \langle c_f^q(X_q),\rho_q\rangle, c_m(X)\}$ is a set of tuples comprised of $q$ FE components and one model component where $\langle c_f^i(X_i),\rho_i\rangle$ represents that the FE component $c_f^i$ will be used in the pipeline with a probability $\rho_i$ and applied on $X_i\subset X$ features in $D$.
\end{definition}

We use the following insights to design our skeleton predictor. First, a pipeline may require several FE components and in many cases the decision of using a particular FE component can be made based on a few meta-features without depending on other FE components. Although occasionally there can be some dependencies between the ML components, our experimental results show that this design decision leads to faster generation of pipelines without sacrificing accuracy. To this end, we design the FE component predictor as a set of binary classifiers $\{\lambda_1, \ldots \lambda_9\}$ that predicts whether a particular FE component $c_f^i \in C_f$ should be used in the generated pipeline for the target dataset, $D$. On the other hand, since by design \tool{} allows only one model $c_m \in C_m$ in a skeleton, we cast the model selection problem as a ranking problem and design a learning-to-rank model to rank all the model components for $D$. 

\begin{definition}[Skeleton Predictor]
\label{def:meta-model}
The skeleton predictor is comprised of a set of sub-models, $\Lambda=\{\lambda_1, \ldots \lambda_9,\lambda_m\}$ where each sub-model approximates a function, $\lambda_i : \mathbb{R}^l\xrightarrow{} y'_i$ where $\mathbb{R}^l$ is a set of meta-feature values and $y'_i$ is a probability score of an ML component ({\it meta-target}) appearing in a pipeline for $D$. 
\end{definition}

{\bf Sub-models to predict the FE components.} FE components are generally applied on a sub-set of features of $D$. For example, an \code{<FE:Imputer>} is generally applied on the features with missing values. Therefore, \tool{} aims to predict not only an FE component ($c^i_f$) for $D$ but also infers the subset of features $X' \subset X$ in $D$ on which $c_f^i$ would be instantiated on. To facilitate the determination of $X'$ for $c^i_f$, a \code{DecisionTreeclassifier} is a natural fit since the classifier would learn a set of precise conditions w.r.t. the meta-features to select  $c^i_f$ for $D$ and later we can analyze those conditions to infer $X'$ for $c^i_f$. However, \code{DecisionTree} models tend to overfit with a large number of features~\cite{scikit-dt}. To minimize the effect of irrelevant features, we first perform a \textit{point biserial correlation} analysis between the meta-features ($\Phi$) and each FE component ($c^i_f$) and use only the meta-features ($\Phi'\subset \Phi$) that exceeds a certain correlation threshold to train $\lambda_i$. 
As a result, we get a set of binary classifiers $\{\lambda_1, \ldots \lambda_9\}$ as the FE components predictor where $\lambda_i$ is used to predict $c^i_f$.

{\bf Sub-model to rank MODEL components.} Unlike the prediction of FE components, which often depends on a few meta-features, it is challenging to determine a few meta-features that can predict the performance of a particular model on $D$~\cite{OBOE-KDD2019}. Therefore, instead of predicting one model based on a few meta-features, we design a learning-to-rank sub-model that considers all the meta-features $\Phi$ to rank all the model components in our corpus. Considering the size of our meta-training dataset, which is not very large and the fact that the ensemble models are better than a single model for complex learning task~\cite{barai1999ensemble}, we designed an ensemble of {\tt LogisticRegression} and {\tt SupportVectorMachine} to rank the model components. More specifically, these meta-models first compute a probability score for each model component and use the average score to sort the target model components.

\subsubsection{Training the Skeleton Predictor (Offline)} 

We trained all the sub-models in the skeleton predictor using the meta-training corpus. Since the proportion of pipelines having and not having a ML component is not equal in the corpus, we used {balanced} weighting strategy to solve the {\it class imbalance} problem. Further, we tuned the hyper-parameters through 5-fold cross validation and grid-search. 

\subsubsection{Generation of Pipeline Skeletons (Online)}

During pipeline generation, \tool{} first computes the set of meta-features, $\Phi$ from $D$ and passes it to the skeleton predictor ($\Lambda$), which returns a set of plausible FE components $\{c_f^1\ldots c_f^q\}$ with probability scores and a ranked-list of the model components $C'_m=[c_m^1,\ldots c_m^r]$. 

{\bf Inferring relevant features.} For each $c_f^i$ in the predicted set, \tool{} infers the relevant features, $X'_i\subset X$ in $D$ on which $c_{f}^i$ can be successfully applied and create the semi-instantiated set of FE components, $C'_f=\{\langle c_f^1(X_1),\rho_1\rangle, \ldots, \langle c_f^q(X_q),\rho_q\rangle\}$. Such inference is important to help avoid pipeline failures caused by infeasible transforms, such as \code{StringVectorizer} applied to a numeric column. Further, it helps precisely identify the columns most suitable for the FE transform, \eg applying \code{SimpleImputer} to only those columns that have missing values.

To infer relevant features for $c_{f}^i$, first \tool{} accesses the the sub-model $\lambda_i$, which is a decision tree classifier that predicted $c_f^i$ for $D$. 
Then \tool{} extracts the decision path that led to the prediction. A decision path is a list of conditions in form of $[\alpha_1~op~v_1, \ldots \alpha_u~op~v_u]$ where $\alpha$, $op$, and $v$ correspond to a meta-feature, $>=$ or $<$, and a real number respectively. Then \tool{} iterates over each feature $x_i\in X$ and selects $x_i$ only if it satisfies at least one of the conditions in the decision path. For example, \tool{} correctly applies the \code{OrdinalEncoder} on \code{card4}, which is a string categorical feature whereas it marks the \code{TransactionAmt} feature as irrelevant, which is indeed a numeric feature. 

\textbf{Generation.} Finally \tool{} selects Top-$k$ models from $C'_m$ and adds one-by-one to the selected FE components to generate $k$ number of pipeline skeletons, $S=[\{C'_f \cup c_m^1\},\ldots \{C'_f \cup c_m^k\}]$.

\subsection{Stage 2: Pipeline Instantiation}
\label{sec:pipeline-instantiation}

This stage synthesizes a set of concrete  pipelines for the given user dataset ($D$) and its predictive task. Given a ranked-list $\mathcal{S}$ of pipeline skeletons produced by pipeline seeding (Section~\ref{sec:pipeline-seeding}), this stage instantiates each skeleton $S$ into a concrete pipeline $P$ by
first creating an ordered skeleton $S_O$ representing a syntactically viable data flow, and then instantiating the components in $S_O$ into a pipeline template, along with necessary glue code. This yields a set of $k$ candidate pipelines, $\mathcal{P}_{cand} = \{ P_1, P_2, \ldots, P_k \}$.

\subsubsection{Create Ordered Skeleton}
The goal of this step is to order the components of $S$, and discard \textit{incompatible components}, if any, to produce an \textit{ordered skeleton} $S_O$, such as the one shown in \Cref{fig:ordered_skeleton}. 
This operation uses a pipeline \emph{dataflow meta-model} extracted by \tool{}, from the meta-learning corpus, during the offline phase. We develop the description using the following terminology.

\begin{definition}[Dataflow dependence]
\label{def:dataflowDep}
There exists a dataflow dependence between components $c_i$ and $c_j$ of a pipeline $P$ for a dataset $D$ iff there exists feature $x_i$ of $D$ on which both $c_i$ and $c_j$ are applied in $P$.
\end{definition}

There exists a dataflow from $c_i$ to $c_j$ in $P$, denoted $c_i \overset{P}{\rightarrow} c_j$,  iff  there is a dataflow dependence between $c_i$ and $c_j$ and $c_i$ precedes $c_j$ in $P$. Dataflow dependence, as defined above, can be inferred through a simple static analysis. The details are elided for brevity. Although neither sound nor complete, this definition provides a simple, efficient way to capture dataflow in all but the most complicated pipelines.

The dataflow meta-model is a partial-order relation $\Delta$ capturing the dataflow between pipeline components, as observed in the corpus pipelines. Specifically,  
\[
   \Delta = \{ (c_i, c_j) \in \mathcal{C} \times \mathcal{C} \mid \exists P \in \mathcal{P}_L, c_i \overset{P}{\rightarrow} c_j \text{ and } \nexists P' \in \mathcal{P}_L, c_j \overset{P'}{\rightarrow} c_i  \}  
\]

The dataflow meta-model is represented as a directed acyclic graph (DAG), $\mathcal{G}_{\Delta}$, whose nodes are the components $\mathcal{C}$ and directed edges $(c_i, c_j) \in \Delta$.

A skeleton $S$ produced by pipeline seeding is transformed into an ordered skeleton $S_O$ using the following steps. First, all dataflow dependencies are captured between potential skeleton components, using Definition~\ref{def:dataflowDep}. If there is any component pair $c_i, c_j$ that has a dataflow dependence but no edge between $c_i, c_j$ in $\mathcal{G}_{\Delta}$, this indicates an incompatible component. Hence the component with the lower predicted probability is discarded. In our motivating example (\Cref{fig:example}), components \code{OneHotEncoder} and \code{OrdinalEncoder}, which happen to be semantic substitutes of each other (\eg convert categorical columns to numeric) present such an instance. Hence, the lower probability component \code{OneHotEncoder} is discarded. Discarding all such components yields a reduced skeleton $S'$.

Next, a sub-graph of the dataflow meta-model $\mathcal{G}_{\Delta}$ with only the nodes in the reduced skeleton $S'$ is extracted. Finally, a topological sort on this sub-graph provides a component order for the reduced skeleton $S'$ consistent with $\mathcal{G}_{\Delta}$. This order is to create the ordered skeleton $S_O$. For our motivating example, \code{Imputer} precedes \code{OrdinalEncoder} in  $S_O$. Reversing the order for a column with missing values would result in a pipeline crash.

\subsubsection{Generate concrete pipeline}
Each ordered skeleton $S_O$ is converted into a concrete pipeline $P$ by instantiating each component $c \in S_O$ in order, into a pipeline template of the kind shown in \Cref{fig:IEE_CIS_pipeline}. Specifically, each $c \in \mathcal{C}_{f} \cup c_{m}$ is instantiated using a parameterized snippet drawn from a small pre-compiled library of standard component implementations, by appropriately filling the parameter holes. For example, \code{OrdinalEncoder} is instantiated by filling the columns hole with \targetColumns (`ProductCD', `card4', $\dots$). \tool{} can also handle type-based instantiation of components. For example, \code{SimpleImputer} is instantiated differently for filling missing values in numeric vs. string columns. 

\subsection{Stage 3: Pipeline Validation}
\label{sec:pipeline-validation}

Each candidate pipeline $P \in \mathcal{P}_{cand}$ is dynamically evaluated to compute an accuracy score (F1/$R^2$), to find the best pipeline, $P_\textit{best}$. \tool{} internally splits the \textit{user-provided} training data $D_{train}$ into {\it internal} training and validation sets which are used for the training and validation of candidate pipelines within this stage. Therefore, the held out \textit{test} dataset, $D_{test}$ (shown as ``test.csv" in \Cref{fig:IEE_CIS_pipeline}) is completely unseen to \tool{}. Finally, $P_\textit{best}$ is used to train on $D_{train}$ and evaluated on $D_{test}$ for the accuracy score returned to the user. 
\section{Evaluation}
\label{sec:evaluation}

Our evaluation addresses the following research questions:
\begin{itemize}
    \item[\textbf{RQ1:}] How does \tool perform compared to the existing state-of-the-art techniques? 
    \item[\textbf{RQ2:}] How robust \tool{} is in producing good quality pipelines across trials?
    \item[\textbf{RQ3:}] Does \tool{} use its search space well to predict a diverse set of FE and model components?
    \item[\textbf{RQ4:}] Does each of the novel technology components of \tool contribute to its effectiveness?
\end{itemize}

\subsection{Experimental Set-up}

\subsubsection{Implementation}
\label{sec:implementation}

\tool{} is implemented in the Python programming language using approximately 5,000 lines of code.   It includes a crawler to download ML projects, a set of tools for required static and dynamic analysis such as mining the order of ML components from a corpus, denoising pipelines, a meta-feature extractor, and machine learning models for the skeleton predictor. \tool uses Kaggle Public APIs~\cite{kaggleapi} for automatically downloading data, the Python-PL library~\cite{pythonpl} to instrument source code for dynamic program slicing, the scikit-learn~\cite{scikit-learn-paper, scikit-learn} library to implement meta-models for the skeleton predictor, and LibCST~\cite{libCST} for static analysis. \tool uses Pandas, Numpy, and Scipy for computing meta-features and its own data analysis.

\subsubsection{Benchmarks}
\label{sec:benchmarks}

We use a set of \numBM{} benchmark datasets to evaluate \tool. This includes the set of 31 datasets used in AL\cite{AL:OOPSLA2019}. They include 12 datasets from the OpenML suite, 9 from PMLB, 4 from Mulan, and 6 from Kaggle.  However, since most of these datasets are small and simple in nature, we have added \numNewBM{} new datasets from Kaggle as representatives of large, real-world datasets which modern AutoML tools should handle. To systematically select the \numNewBM{} new benchmark datasets we collected all the {\it Featured} and {\it Playground} Kaggle competitions completed since the year 2015. From these we selected ones operating only on tabular data and where the license permits academic research and use outside the competition. Finally, we selected \numNewBM{} datasets based on size either in terms of large number of rows or columns, or have various types of columns. Table~\ref{tbl:effectiveness} presents the size, prediction task, and source repository for each benchmark dataset.

\subsubsection{Experimental Methodology}

\tool{} is trained on our meta-learning corpus of \numCorpNtbks{} pipelines and corresponding cleaned datasets. Therefore, the 41 benchmark datasets are completely {\it unseen} to \tool{}. Similar to AL~\cite{AL:OOPSLA2019} and auto-sklearn~\cite{Auto-sklearn-NIPS2015}, we performed 10 trials of each experiment for each benchmark with a one hour time out. For each trial, we randomly split the \textit{user-provided} dataset into \textit{training} and \textit{testing} data in a 75:25 split. Then \tool{} generated a pipeline using \textit{only} the \textit{user-provided} training data and then reported its accuracy on the \textit{user-provided} testing data.
All the baselines and existing tools were run using the same train-test split of data in each trial to ensure a fair comparison. We use standard {\it macro} F1 scores and $R^2$ scores for classification and regression tasks respectively, and used the mean score of 10 runs to compare the results, following existing literature~\cite{Auto-sklearn-NIPS2015, AL:OOPSLA2019}.
We ran all tools on 4 vCPUs of Xeon E5-2697A v4 (2.60GHz) with 16GB memory for \openml, PMLB, and \mulan datasets and on 8 vCPUs with 32GB memory for Kaggle datasets.

\subsection{RQ1: \tool{} versus state of the art}

\begin{table*}[t]
  \centering
  \footnotesize
  \caption{Effectiveness of \tool compared to the state-of-the-art AutoML tools on the benchmark datasets. Bold numbers indicate the best scores; Underlined numbers are not statistically different from the best scores according to a Wilcoxon-signed-rank Test ($\alpha=0.05$). {\it Failed} means the tool has failed on all the 10 trials whereas {\it F[n]} means the tool has failed on $n$ trials.}
  \label{tbl:effectiveness}
\begin{tabular}{lrrrrrrrrrrrr}
\hline
Dataset & \tool & AL & auto-sk. & TPOT & Basic ML & Default & Base. 1 & Base. 2 & Metric & Source & Rows & Cols\\ \hline
1049 & \underline{0.78} & 0.73 & \textbf{\underline{0.79}} & \underline{0.78} & 0.63 & 0.47 & \underline{0.78} & 0.64 & F1 & OpenML & 1458 & 37 \\
1120 & \underline{0.87} & 0.83 & \textbf{\underline{0.87}} & 0.86 & 0.75 & 0.39 & F[5] & 0.75 & F1 & OpenML & 19020 & 10 \\
1128 & 0.95 & 0.94 & \textbf{\underline{0.97}} & 0.96 & 0.94 & 0.44 & F[5] & 0.94 & F1 & OpenML & 1545 & 10935 \\
179* & \textbf{\underline{0.79}} & 0.77 & Failed & Failed & 0.43 & 0.43 & F[7] & 0.79 & F1 & OpenML & 48842 & 14 \\
184 & \textbf{\underline{0.77}} & 0.55 & Failed & Failed & Failed & 0.02 & 0.67 & 0.37 & F1 & OpenML & 28056 & 6 \\
293* & \textbf{\underline{0.96}} & 0.78 & 0.91 & 0.75 & 0.75 & 0.34 & \underline{0.96} & 0.75 & F1 & OpenML & 581012 & 54 \\
38 & \textbf{\underline{0.95}} & \underline{0.94} & Failed & Failed & 0.57 & 0.48 & 0.87 & 0.83 & F1 & OpenML & 3772 & 29 \\
389 & \textbf{\underline{0.83}} & 0.63 & \underline{0.83} & 0.79 & 0.81 & 0.02 & 0.83 & 0.81 & F1 & OpenML & 2463 & 2000 \\
46 & \underline{0.95} & \underline{0.95} & Failed & Failed & Failed & 0.23 & \textbf{\underline{0.96}} & 0.94 & F1 & OpenML & 3190 & 60 \\
554 & 0.98 & 0.91 & \textbf{\underline{0.98}} & 0.83 & 0.92 & 0.02 & 0.98 & 0.92 & F1 & OpenML & 70000 & 784 \\
772 & \underline{0.51} & \underline{0.51} & \textbf{\underline{0.51}} & \underline{0.51} & 0.41 & 0.35 & \underline{0.51} & \underline{0.49} & F1 & OpenML & 2178 & 3 \\
917 & 0.90 & \underline{0.90} & \underline{0.90} & \textbf{\underline{0.90}} & 0.65 & 0.35 & \underline{0.90} & 0.65 & F1 & OpenML & 1000 & 25 \\
Hill\_Valley\_with\_noise & 0.95 & 0.73 & \textbf{\underline{1.00}} & 0.99 & 0.95 & 0.33 & 0.95 & 0.95 & F1 & PMLB & 1212 & 100 \\
Hill\_Valley\_without\_noise & 0.99 & 0.69 & \textbf{\underline{1.00}} & \underline{1.00} & \underline{1.00} & 0.32 & \underline{1.00} & \underline{1.00} & F1 & PMLB & 1212 & 100 \\
breast\_cancer\_wisconsin & \textbf{\underline{0.97}} & 0.95 & \underline{0.97} & \underline{0.96} & 0.93 & 0.38 & \underline{0.97} & 0.93 & F1 & PMLB & 569 & 30 \\
car\_evaluation & 0.95 & 0.97 & \underline{0.98} & \textbf{\underline{0.99}} & 0.76 & 0.21 & 0.95 & 0.79 & F1 & PMLB & 1728 & 21 \\
glass & \textbf{\underline{0.74}} & \underline{0.67} & 0.63 & \underline{0.70} & 0.48 & 0.10 & 0.71 & 0.45 & F1 & PMLB & 205 & 9 \\
ionosphere & \textbf{\underline{0.94}} & 0.91 & \underline{0.94} & \underline{0.94} & 0.84 & 0.39 & \underline{0.94} & 0.86 & F1 & PMLB & 351 & 34 \\
spambase & \underline{0.96} & 0.94 & \underline{0.95} & \underline{0.95} & 0.92 & 0.38 & \textbf{\underline{0.96}} & 0.94 & F1 & PMLB & 4601 & 57 \\
wine\_quality\_red & \textbf{\underline{0.35}} & \underline{0.33} & \underline{0.33} & \underline{0.34} & 0.22 & 0.10 & \underline{0.34} & 0.29 & F1 & PMLB & 1599 & 11 \\
wine\_quality\_white & \textbf{\underline{0.44}} & 0.42 & 0.41 & \underline{0.43} & 0.16 & 0.10 & \underline{0.44} & 0.34 & F1 & PMLB & 4898 & 11 \\
detecting-insults-in-social-comm... & 0.71 & \underline{0.76} & Failed & Failed & \textbf{\underline{0.77}} & 0.42 & Failed & 0.42 & F1 & Kaggle & 3947 & 2 \\
housing-prices & \textbf{\underline{0.89}} & 0.85 & Failed & Failed & 0.80 & -0.00 & Failed & F[6] & R2 & Kaggle & 1460 & 80 \\
mercedes-benz & \underline{0.52} & \textbf{\underline{0.53}} & Failed & Failed & -2.0E+23 & -0.00 & -0.80 & Failed & R2 & Kaggle & 4209 & 377 \\
sentiment-analysis-on-movie-rev...* & \textbf{\underline{0.49}} & 0.39 & Failed & Failed & F[7] & 0.13 & Failed & 0.02 & F1 & Kaggle & 156060 & 3 \\
spooky-author-identification & 0.78 & 0.80 & Failed & Failed & \textbf{\underline{0.81}} & 0.19 & 0.78 & 0.19 & F1 & Kaggle & 19579 & 2 \\
titanic & 0.79 & 0.71 & Failed & Failed & \textbf{\underline{0.81}} & 0.38 & Failed & 0.79 & F1 & Kaggle & 891 & 11 \\
enb & 0.98 & 0.98 & \textbf{\underline{0.99}} & Failed & 0.89 & -0.01 & 0.98 & 0.96 & R2 & Mulan & 768 & 8 \\
jura & 0.60 & \textbf{\underline{0.76}} & 0.48 & Failed & 0.52 & -0.01 & 0.59 & 0.60 & R2 & Mulan & 359 & 15 \\
sf1 & \underline{-0.09} & F[4] & Failed & Failed & Failed & \textbf{\underline{-0.01}} & \underline{-0.10} & \underline{-0.05} & R2 & Mulan & 323 & 10 \\
sf2 & \textbf{\underline{0.05}} & F[3] & Failed & Failed & Failed & -0.00 & -1.0E+23 & -4.6E+22 & R2 & Mulan & 1066 & 10 \\
costa-rica* & 0.91 & 0.87 & Failed & Failed & 0.21 & 0.19 & \textbf{\underline{0.92}} & 0.55 & F1 & Kaggle & 9557 & 142 \\
Categorical-Feature-Enc...-Chal...-II* & \textbf{\underline{0.58}} & 0.54 & Failed & Failed & Failed & 0.45 & Failed & Failed & F1 & Kaggle & 600000 & 24 \\
Porto-Seguros-Safe-Driver-Pred... & 0.49 & F[2] & \textbf{\underline{0.52}} & 0.52 & 0.49 & 0.49 & 0.49 & 0.49 & F1 & Kaggle & 595212 & 58 \\
Kobe-Bryant-Shot-Selection* & \textbf{\underline{0.64}} & Failed & Failed & Failed & 0.36 & 0.36 & Failed & 0.62 & F1 & Kaggle & 30697 & 24 \\
whats-cooking & \underline{0.71} & Failed & Failed & Failed & \textbf{\underline{0.71}} & 0.02 & \underline{0.71} & 0.02 & F1 & Kaggle & 39774 & 2 \\
PUBG-Finish-Placement-Prediction* & \textbf{\underline{0.93}} & Failed & -0.00 & 0.86 & 0.83 & -0.00 & 0.93 & 0.83 & R2 & Kaggle & 4446966 & 28 \\
Santander-Value-Prediction-Chal... & 0.12 & Failed & \underline{0.26} & \textbf{\underline{0.27}} & -7.2E+17 & -0.00 & 0.12 & Failed & R2 & Kaggle & 4459 & 4993 \\
IEEE-CIS-Fraud-Detection* & \textbf{\underline{0.82}} & Failed & Failed & Failed & 0.49 & 0.49 & Failed & Failed & F1 & Kaggle & 590540 & 394 \\
Quora-Insincere-Questions-Class...* & \textbf{\underline{0.75}} & 0.63 & Failed & Failed & Failed & 0.48 & Failed & 0.06 & F1 & Kaggle & 1306122 & 3 \\
DonorsChoose.org-App...-Screening* & \textbf{\underline{0.50}} & F[5] & Failed & Failed & Failed & 0.46 & Failed & Failed & F1 & Kaggle & 182080 & 16 \\
\hline
\#champions & 19 & 2 & 9 & 3 & 4 & 1 & 3 & 0 \\
\#winners & 27 & 9 & 17 & 12 & 5 & 1 & 14 & 3 \\
\#failures & 0 & 9 & 19 & 21 & 8 & 0 & 12 & 6 \\
\hline
\end{tabular}
\end{table*}

We compared the performance of \tool{} to three state-of-the-art AutoML systems: \autosk~\cite{Auto-sklearn-NIPS2015} (ver. 0.12.2), TPOT~\cite{TPOT:GECCO2016} (ver. 0.11.7), and AL~\cite{AL:OOPSLA2019}, from its public distribution \cite{AL-github} using the same configurations as in \cite{AL:OOPSLA2019}. \autosk is an actively managed open-source project on Github with more than 6K stars. AL represents the most recent AutoML technique that also learns from human-written pipelines to generate supervised pipelines.
In addition, we implemented two baseline tools \basicml and \default, representing basic ML techniques, following the methodology described in \cite{AL:OOPSLA2019}. 
Specifically, \basicml applies \code{SimpleImputer} to fill numeric and string missing values with 0 and empty string respectively, \code{CountVectorizer} to transform all string columns to token counts, and then uses the \code{LogisticRegression} and \code{LinearRegression} models for classification and regression tasks respectively. \default always predicts the most frequent label for classification tasks or the mean value for regression tasks.

\subsubsection{Quantitative Comparison} Table~\ref{tbl:effectiveness} presents the evaluation results in terms of average macro F1 and $R^2$ scores over 10 runs for classification and regression tasks respectively. Highest scores for each benchmark are marked as bold. We call them as {\it champion}. Furthermore, we performed a pair-wise Wilcoxon-signed-rank Test ($\alpha=0.05$) to see whether the score difference between the champion and another tool for a benchmark is statistically significant across 10 trials. The underlined numbers represent the scores that are statistically similar to the champion. We call them as {\it winners}.  

We start by comparing the two baseline tools: \basicml and \default to all other tools. As expected, \default's simplistic prediction performed the worst. Interestingly, \basicml is the champion on 4 datasets, since some of the datasets are simple and do not need any sophisticated pipelines. However, \basicml's overall performance is poor compared to any other AutoML tools, in terms of mean F1/R2 scores. Therefore, this result  supports the {\it no free lunch} hypothesis~\cite{wolpert1997no} that no single pipeline is good for every dataset. 

 Comparing the performance of \tool{} to other AutoML tools, Table~\ref{tbl:effectiveness} shows that \tool outperforms the state-of-the-art AutoML tools in terms of successful pipeline generation, number of champions, and winners. \tool{} generated a successful pipeline for each benchmark and trial, whereas AL, \autosk, and TPOT failed on 9, 17, and 12 datasets respectively. There are several reasons for failures including not being able to handle various types of data, unexpected exceptions, applying FE components on inappropriate columns, or timeout.

In terms of performance score, \tool{} is champion for 18 subjects, whereas the second best tool, AL, based on the number of successful pipelines, is champion for only 2 datasets. On the other hand, although \autosk failed on highest number of datasets, it is champion for 9 datasets. These results indicate that \autosk performs well in a limited scope. However, although AL has a broader scope, it has overall performed moderately. Interestingly, \tool{} outperforms them both in terms of scope and performance. The same findings also hold in terms of number of winners. \tool performed the best or comparable to the best for \numWinOrComparable{} datasets, which is the highest among all tools.  

For the 10 more difficult datasets (marked with a * in Table~\ref{tbl:effectiveness}) -- the largest ($row \times columns$) datasets requiring at least one FE component -- \tool performs even better relative to other tools. 
\tool{} produces best or comparable performance on \numLargeSMLWins\ of them, with AL failing to produce a pipeline on 4 of them and TPOT, \autosk on most of them. These results illustrate the value of \tool{}'s 
divide-and-conquer synthesis to produce viable pipelines especially for large, complex datasets.

\subsubsection{Qualitative Analysis} 
 
We analyze the results qualitatively using a few concrete examples. Benchmark {\tt OpenML-293} presents an interesting case where every tool produced a pipeline since the dataset contains only numeric values. AL predicted and selected XGBoostClassifier through dynamic evaluation, which achieved a macro mean F1 score of 0.78. AL's prediction may suffer since it uses language model which depends on the previous two components for prediction. However, there is no need for FE components for this dataset. \autosk selected a pipeline based on dataset similarity that performs standard scaling first and then uses \code{GradientBoostingClassifier}. It achieved an F1 score of 0.91, better than AL. However, \tool{} predicted an even better model: \code{RandomForestClassifier}, which achieved the best F1 score:~0.96. 

For the {\it sentiment-analysis-on-movie-reviews} dataset,  \autosk simply failed since it cannot handle textual data. In contrast, AL and \tool{} both successfully generated pipelines by using a \code{TextVectorizer} component to convert text to numeric columns. However, AL selected the \code{LinearSVC} model that resulted in an F1 score of 0.39. On the other hand, \tool{} selected an additional text preprocessing components that performs some basic cleaning and normalization of text. Further, it selected a better model \code{CatBoost} that provided an F1 score of 0.49.

Finally, the {\tt jura} dataset presents a negative example for \tool{} where AL achieved a significantly better score. On investigating the reason, we found that AL selected a model called {\tt Ridge}, which is not used in our project corpus. Even for AL, this particular model was not highly ranked by its meta-model. However, since AL set a beam\_size of 30 for this dataset, \ie it evaluated 30 different models to select the best model, AL was successful in this case. AL could afford to perform such extensive evaluation for this dataset simply because the dataset is small and hence the dynamic evaluation is fast. However, for this extensive dynamic evaluation, AL failed to produce any pipeline for large datasets such as {\tt IEEE} and {\tt Donors} due to timeout. In contrast, \tool{} uses only top 3 models based on its meta-model and performs overall the best.

\subsection{RQ2: Robustness of \tool{}}

\begin{figure}
\captionsetup{justification=centering}
\centering
    \includegraphics[width=0.9\columnwidth]{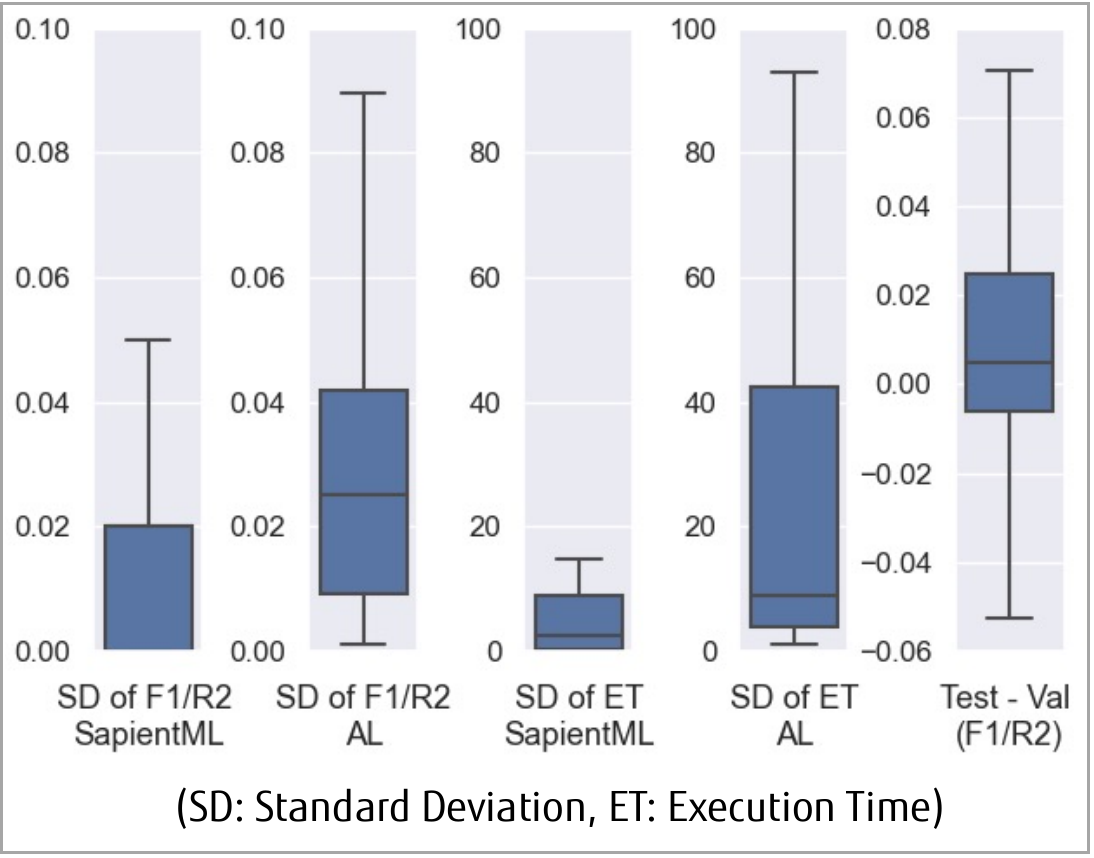}
    \caption{\label{fig:robustness}Robustness of \tool{}}
\end{figure}

We analyze the robustness of \tool{} in generating pipelines in terms of variation of performance scores and execution time across 10 trials. 
To investigate how much \tool{} fluctuates across trials, we calculated the standard deviation of performance scores and execution time across 10 trials for each benchmark dataset. Figure~\ref{fig:robustness} presents the distribution of the standard deviations across 10 trials for 41 benchmark datasets. The results show that \tool{} is overall very stable across the 10 trials in terms of both accuracy and execution time. Both the 50th percentile (mean) and 75th percentile standard deviation for macro F1/R2 scores across all the benchmark are only 0.02, which is more stable than that of AL, which are 0.03 and 0.04 respectively. The same is true for the execution time. The 50th percentile and 75th percentile standard deviation of execution time are only 3 and 9 seconds respectively for \tool{}, which are 9 and 41 seconds respectively for AL.

Finally, we investigate whether the generated pipelines are overfitted to the corresponding training data. To prevent overfitting, we already made sure that \tool{} generates pipeline only using 75\% data, and the generated pipeline is tested on 25\% {\it held-out} test data. However, in this RQ, we investigate even further. Generally overfitting happens when a model performs very good on the validation data but performs poorly on the test data~\cite{subramanian2013overfitting}. To this end, we compute the internal validation score based on which \tool{} selected the best pipeline. Then we compare the validation accuracy with held-out test accuracy. As the fifth boxplot in Figure~\ref{fig:robustness} shows, the 50th and 75th percentile difference between test and validation accuracy are 0.01 and 0.02 respectively.  And interestingly, the differences are positive, which means that the final test scores are better than the validation scores for most of the subjects. We could not compare this result with any other tools since we do not have access to their validation scores. 

\subsection{RQ3: Diversity in Meta-Prediction}

\begin{figure}
    \includegraphics[width=\columnwidth]{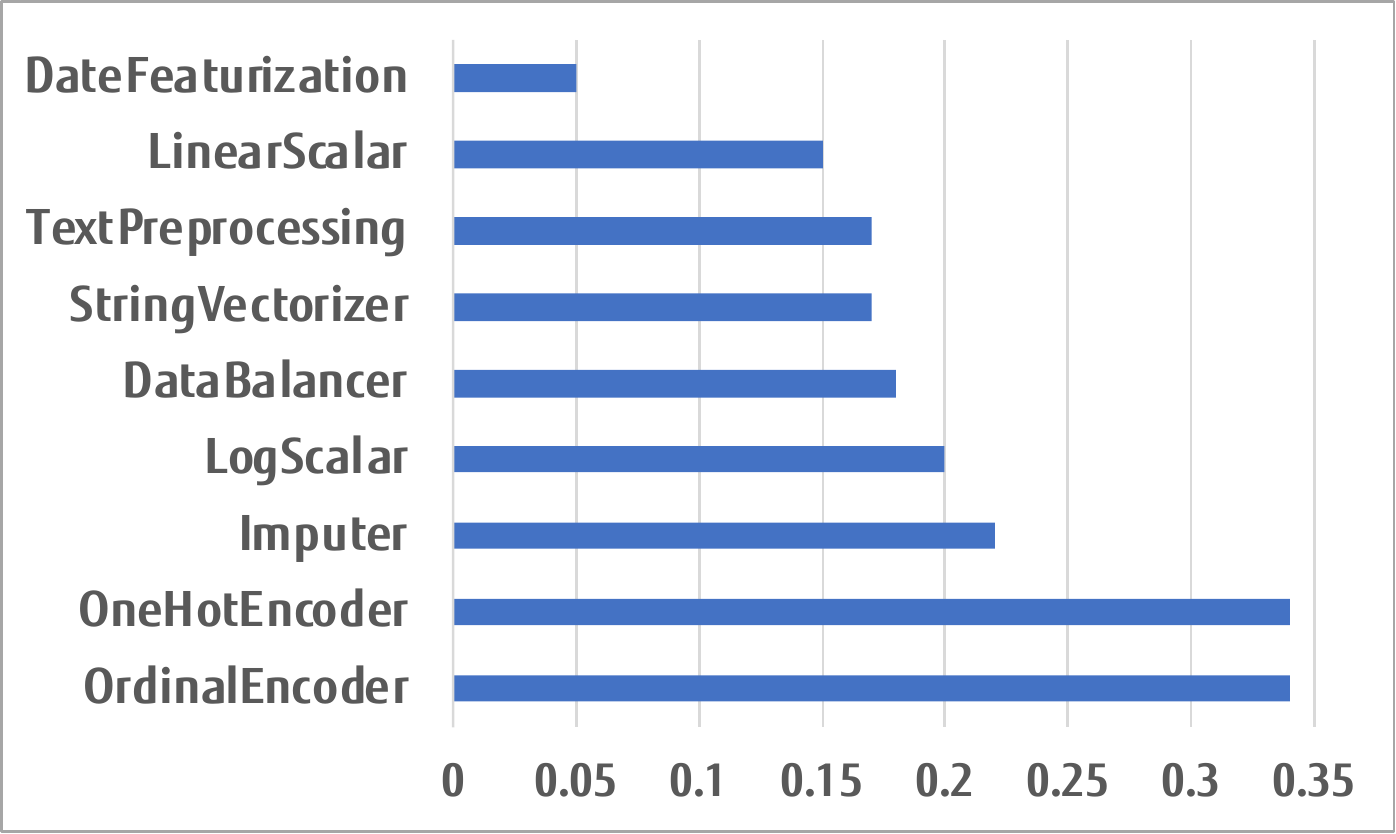}
    \vspace{-18pt}
    \caption{\label{fig:p-comp-dist}Distribution of FE predictions}
\end{figure}

\begin{figure}
    \includegraphics[width=\columnwidth]{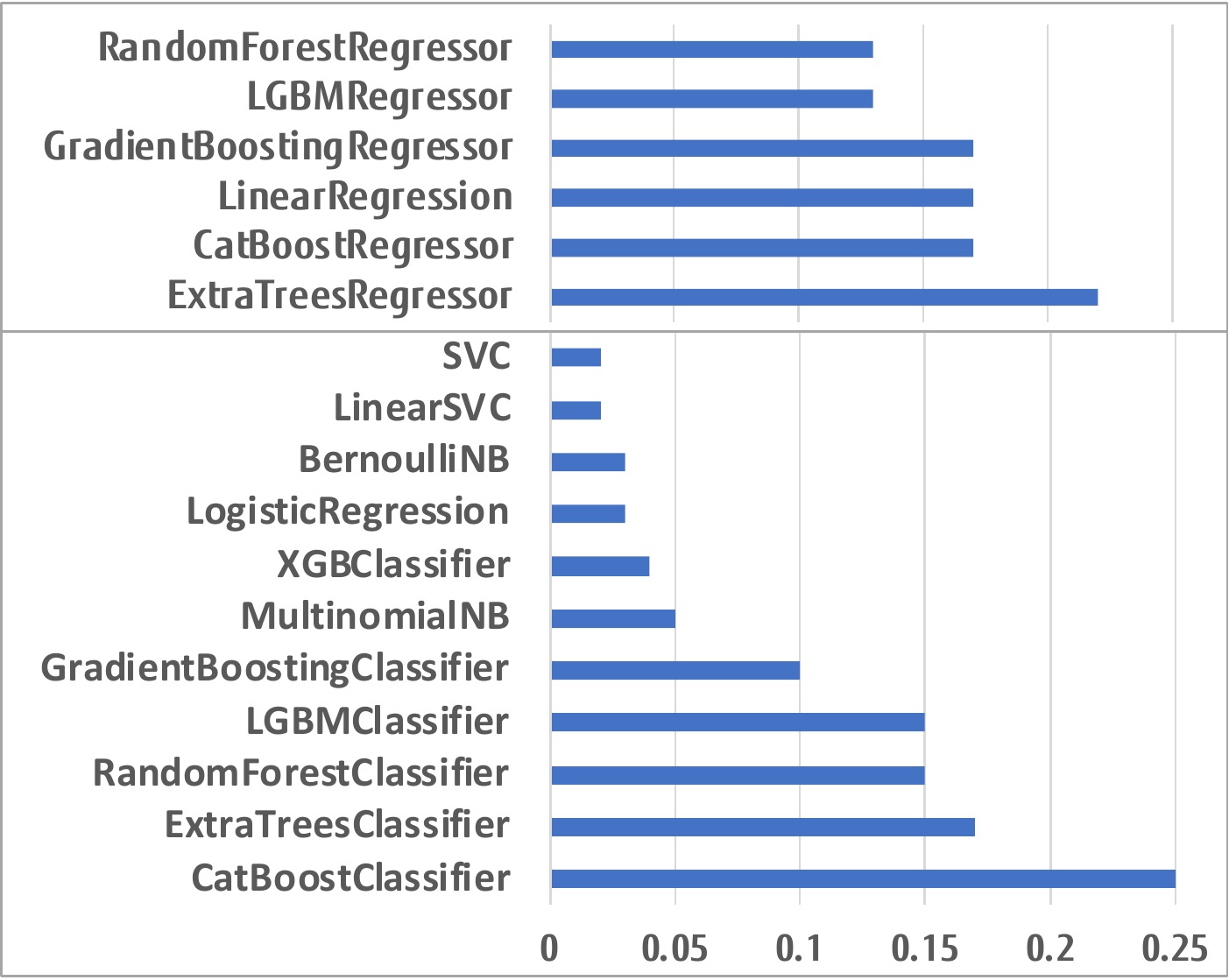}
    \vspace{-18pt}
    \caption{\label{fig:m-comp-dist}Distribution of model predictions}
\end{figure}

Figure~\ref{fig:p-comp-dist} presents the distribution of predicted ML components in pipelines skeletons for all benchmarks and trials.  The results show that the skeleton predictor was able to predict all the 9 FE components successfully. Among these components, accurate prediction of \code{Imputer}, \code{OrdinalEncoder} or \code{OneHotEncoder}, \code{TextVectorizer}, and  \code{DateFeaturization} is important since any false negative predictions may lead to a crash during pipeline execution. Since \tool{} is able to produce a successful pipeline for each trial in each benchmark dataset (Table~\ref{tbl:effectiveness}), it is evident that the skeleton predictor accurately predicted these components. Similarly, as Figure~\ref{fig:m-comp-dist} shows, the skeleton predictor is able to predict a wide range of model components. More specifically, it predicted 11 different classification and 6 regression models for 33 classification and 8 regression tasks respectively. As expected, some models such as \code{CatBoost} and \code{RandomForest} are dominant since they are fundamentally better. However, traditional models such as \code{LogisticRegression} or \code{SVC} are also predicted depending on the dataset. The overall results suggest that the predictions were effective for most of the datasets. 

\subsection{RQ4: Impact of novel components}
This research question investigates the contribution of \tool's two main components: i) pipeline seeding and ii) pipeline instantiation. To this end, we create two baselines: 

{\bf \baseadapt.} This baseline uses the skeleton predicted by pipeline seeding but instantites each FE component on the entire dataset.

{\bf \baseseeding.} In this baseline, we further relax \baseadapt by replacing the pipeline seeding by a common skeleton to understand the combined effect of pipeline seeding and instantiation. To create the default skeletons, we take three most frequently used FE components in our corpus, one at a time, with the most frequent model. Thus, we try with three skeletons and take the best accuracy. 

Table~\ref{tbl:effectiveness} shows the results of  \baseadapt and \baseseeding (columns 8 and 9). From the results, it can be seen that the two baselines fail on 6 and 12 datasets, respectively. Further, their performance is poor due to the use of FE components on all columns in the dataset. \baseadapt achieves comparable performance to \tool for datasets that are simple and do not require any FE components. However, the overall results show that both pipeline seeding and pipeline instantiation are important for \tool to succeed.

\section{Limitations \& Threats to Validity}
\label{sec:limitations}

\textbf{External validity.} Our framework has only been instantiated for ML pipelines in Python and evaluated on our \numBM\ benchmark datasets. Thus our results may not hold outside this scope. We tried to mitigate this risk by using standard benchmarks from previous work~\cite{AL:OOPSLA2019, Auto-sklearn-NIPS2015, TPOT:GECCO2016} augmented with large, diverse, real-world datasets used on public data science competitions hosted on Kaggle.

\textbf{Quality of data.} Being a data-driven technique \tool's performance is inherently limited by the quality of its training data. It is a well known problem that most notebooks available on Kaggle or \gh cannot be locally re-executed~\cite{Wang:ASE2020, Wang:FSE2020}. Thus, we could also mine only a fraction of the data (\ie pipelines) potentially available on Kaggle. Further, the notebooks we did obtain vary significantly in quality and their use of specific libraries versus custom code. These differences manifest as noise in our analysis. We tried to mitigate these issues by developing simple but effective corpus augmentation (Section~\ref{sec:corpus-augmentation}), pipeline denoising (Section~\ref{sec:denoise-pipelines}) and by using semantic components classes (Section~\ref{sec:pipeline-seeding}) to canonicalize pipelines. However, using a larger, cleaner data corpus could significantly strengthen our results. 

\textbf{Simple skeleton predictor model.} Currently, our skeleton predictor uses a rather simple model that prioritizes features of the dataset and ignores correlations between (predicted) pipeline components. This approximation allows the model to perform well with limited data, as it did on our benchmarks. However, generating much more deeper or sophisticated pipelines might necessitate a more expressive model trained on substantially larger, cleaner data.

\textbf{Manual definition of the pipeline space.} Currently, we use a manual methodology to define the synthesis space of \tool, including creating the clusters of APIs constituting the semantic FE classes (Section~\ref{sec:pipeline-seeding}). We note that this is consistent with the practice of previous AutoML techniques~\cite{TPOT:GECCO2016, Auto-sklearn-NIPS2015, AL:OOPSLA2019}. However, we follow a transparent and systematic process (Section~\ref{sec:pipeline-seeding}), so that \tool can be easily generalized to other ML components once viable pipeline data demonstrating their use is available. However, this would still be limited to API-based ML components. The problem of mining and re-using arbitrary, custom ML transforms in pipeline synthesis remains a very interesting, open problem. 

\textbf{Hyper-parameter optimization (HPO).} \tool focuses on ML component selection and end-to-end pipeline instantiation. HPO is currently out of its scope. However, standard Bayesian optimization HPO~\cite{smac3} could be added as a post-processing step.    
\section{Related Work}
\label{sec:related-work}

\textbf{AutoML for tabular data.} Previous AutoML techniques use different techniques to explore the huge combinatorial search space of potential candidate pipelines.  TPOT~\cite{TPOT:GECCO2016} uses evolutionary search while ReinBo~\cite{xudong2019reinbo} uses Reinforcement Learning combined with Bayesian Optimization~\cite{Hutter2011}. Auto-WEKA~\cite{Auto-WEKA-KDD2013}, and later Auto-Sklearn~\cite{Auto-sklearn-NIPS2015, ASKL2}, employ meta-learning on a corpus of synthetic optimized pipelines to \textit{select} the most appropriate pipeline and then tune hyper-parameters using Bayesian Optimization. TensorOBOE~\cite{TensorOBOE-KDD2020, OBOE-KDD2019} builds on this approach using low rank tensor decomposition as a surrogate model for efficient pipeline search. AL~\cite{AL:OOPSLA2019} uses language models learned from human-written pipelines, in combination with aggressive dynamic evaluation of partial pipelines, to explore the pipeline space. AMS~\cite{AMS:FSE2020} mines constraints from corpora of human-written pipelines to help warm-start search-based AutoML like TPOT. \tool shares AL and AMS's goal of learning from human-written pipelines. However, unlike all of the above approaches, which essentially reason on complete pipelines, \tool combats AutoML combinatorial state space explosion through a novel divide-and-conquer approach of first reasoning on individual ML components and subsequently assembling a small pool of candidate pipelines for final analysis.

\textbf{AutoML for DL models.} This area is reviewed extensively in \cite{automl-survey2018, automl-survey2019}.  This research focuses on synthesizing the neural network models themselves, through \emph{neural architecture search (NAS)}~\cite{NAS1, NAS2, automl_book}, or on hyper-parameter optimization (HPO)~\cite{HPO-chapter, automl_book}.  By contrast, \tool addresses 
ML component selection and end-to-end pipeline instantiation, treating ML components as black-boxes. 

\textbf{Program synthesis for data wrangling.} These techniques typically use input-output examples of data-frames as an input specification to synthesize programs implementing data wrangling operations (data pre-processing, cleaning, transformation) for the given dataset. 
They prune or navigate the synthesis program space by manually specified API constraints coupled with constraint-solving~\cite{Feng-PLDI2017}, automatically learning lemmas during synthesis~\cite{Neo-PLDI2018}, or using more general neural-network-backed program generators~\cite{autopandas2019}. 
However, the PbE paradigm common to these techniques is not applicable to ML pipeline synthesis.

\textbf{ML-based program synthesis.} One class of approaches, such as \cite{Raychev2014, Lee2018}, use probabilistic models trained on programs extracted from large open repositories (\eg Github and StackOverflow) to rank the space of candidate programs generated by the synthesizer. Another body of work~\cite{singh2017ap, Shu2017, murali2018neural, sun2019} leverages user-provided input-output examples, or natural language description, to create a search space for neural program synthesis, typically for simple domains such as string-manipulating programs. By contrast, our synthesis technique is specifically engineered to use a given dataset and its predictive task as the (only) specification for synthesis.

\section{Conclusions}
\label{sec:conclusions}

In this work we proposed a learning-based AutoML technique \tool{}, to generate supervised ML pipelines for tabular data. \tool combats the huge combinatorial search space of AutoML through a novel divide-and-conquer three-stage program synthesis approach that reasons on successively smaller search spaces. We have instantiated \tool\ as part of a fully automated tool-chain that creates a cleaned, labeled learning corpus by mining Kaggle, learns from it, and uses the learned models to then synthesize pipelines for new predictive tasks. We evaluated \tool\ on a set of \numBM\ benchmark datasets and against \numAutoMLComp\ state-of-the-art AutoML tools and \numBaselines\ baselines. Our evaluation showed that \tool{} produced the best or comparable accuracy in \numWinOrComparable\ of the benchmarks while the second best tool failed to even produce a pipeline on \numAlFails\ of the instances. 

\bibliographystyle{ACM-Reference-Format}
\bibliography{references}
 
\end{document}